\documentclass{article}

\usepackage[dblblindworkshop,final]{tagds_2025}

\workshoptitle{Topology, Algebra, and Geometry in Data Science}

\usepackage[utf8]{inputenc} 
\usepackage[T1]{fontenc}    
\usepackage{hyperref}       
\usepackage{url}            
\usepackage{booktabs}       
\usepackage{amsfonts}       
\usepackage{nicefrac}       
\usepackage{microtype}      
\usepackage{xcolor}         

\usepackage{ngkarris-math}
\usepackage{graphicx}
\usepackage{subcaption}

\title{Which Way from B to A: The role of embedding geometry in image interpolation for Stable Diffusion}

\author{
  Nicholas Karris \\
  Department of Mathematics\\
  University of California, San Diego\\
  La Jolla, CA 92093 \\
  \texttt{nkarris@ucsd.edu} \\
  \And
  Luke Durell \\
  National Security Directorate\\
  Pacific Northwest National Laboratory\\
  Richland, WA 99354 \\
  \texttt{luke.durell@pnnl.gov} \\
  \And
  Javier E. Flores \\
  Earth and Biological Sciences Directorate\\
  Pacific Northwest National Laboratory\\
  Richland, WA 99354 \\
  \texttt{javier.flores@pnnl.gov} \\
  \And
  Tegan Emerson \\
  National Security Directorate\\
  Pacific Northwest National Laboratory\\
  Richland, WA 99354 \\
  \texttt{tegan.emerson@pnnl.gov} \\
}

\begin{document}

\maketitle

\begin{abstract}
   It can be shown that Stable Diffusion has a permutation-invariance property with respect to the rows of Contrastive Language-Image Pretraining (CLIP) embedding matrices. This inspired the novel observation that these embeddings can naturally be interpreted as point clouds in a Wasserstein space rather than as matrices in a Euclidean space.
   This perspective opens up new possibilities for understanding the geometry of embedding space. For example, when interpolating between embeddings of two distinct prompts, we propose reframing the interpolation problem as an optimal transport problem.
   By solving this optimal transport problem, we compute a shortest path (or geodesic) between embeddings that captures a more natural and geometrically smooth transition through the embedding space.
   This results in smoother and more coherent intermediate (interpolated) images when rendered by the Stable Diffusion generative model.
   We conduct experiments to investigate this effect, comparing the quality of interpolated images produced using optimal transport to those generated by other standard interpolation methods.
   The novel optimal transport--based approach presented indeed gives smoother image interpolations, suggesting that viewing the embeddings as point clouds (rather than as matrices) better reflects and leverages the geometry of the embedding space.
\end{abstract}

\section{Introduction} \label{sec:intro}

The study of the manipulation and interpolation of the inputs to and outputs of generative diffusion-based text-to-image models like Stable Diffusion and latent diffusion models \citep{rombach_high-resolution_2022, esser_scaling_2024} has attracted increased attention in recent years. These models produce novel images by denoising a random noise latent conditioned on a prompt input \citep{ho_denoising_2020}. The prompt embeddings obtained from the raw prompt text, such as those obtained by CLIP \citep{radford_learning_2021}, are the input observed by the model and used for training and sampling. Understanding these prompt embedding spaces is desirable for a variety of creative and functional applications, including prompt optimization \citep{gal_image_2022, wang_discrete_2024, zhu_image_2007}, improved sampling diversity \citep{deckers_manipulating_2024}, image and prompt inversion \citep{zhang_compositional_2024, li_editor_2025}, concept ablation and unlearning \citep{li_get_2023, kumari_ablating_2023}, and image interpolation \citep{wang_interpolating_2023}.

Specifically, interest in image interpolation in the context of diffusion models has grown. Applications include anticipated domains such as video transitions \citep{zhang_tvg_2025}, but also extend to a variety of unexpected use-cases, such as improving echocardiography image quality \citep{sivaanpu_denoising_2024} or classifying plant health \citep{lee_enhancing_2025}. Methods to obtain smooth image interpolations in the context of generative diffusion models include latent-space interpolation \citep{yu_probability_2025, saito_image_2025} a mixture of latent and prompt embedding interpolation \citep{wang_interpolating_2023, yang_impus_2023, zhang_diffmorpher_2024}, and even attention interpolation \citep{he_aid_2024}. Most of these approaches (excepting \citet{he_aid_2024}) focus on starting with real images, inverting to the embedding space, interpolating between embeddings, and then generating images for the interpolated embeddings.

In parallel, there has been a growing interest in the geometry of learned latent spaces for frontier models like the work of \cite{balestriero2025geometry,lee2023geometry, arvanitidis2017latent,sakamoto2024geometry}, and even for movement and interpolation between images in \cite{park2023understanding}. Geometric perspectives are being leveraged extensively in the the interpretability literature \cite{voynov2020unsupervised,balestriero2020mad}. Additionally, there is research in the broader community looking at how geometric properties are connected to topics like hallucination \cite{yeats2025geometric, phillips2025geometric} and deep-fake \cite{xie2025grdt, barnabo2023deep,sivabalamurugan2024deepfake} detection. Much of this work focuses on clustering, structure, and dominant modes or directions within the learned embedding space where the representations of the data are vectors or matrices. 

Rather than focus on the geometry of \emph{latent} space as in, e.g., \cite{park2023understanding}, we look to connect geometric perspectives on the textual \emph{embedding} space with the image interpolation task in Stable Diffusion.
This novel work leverages a permutation invariance property of the embedding-to-image generator used in Stable Diffusion to produce a new way of exploiting the geometry of embeddings in image interpolation. Rather than understanding relationships between prompts based on Euclidean or matrix-based relationships, we consider learned embeddings as unordered point-clouds. This point-cloud provides a different way to interpolate between embeddings using optimal transport.
While optimal transport techniques have been incorporated into diffusion frameworks for concept optimization \citep{li_optimal_2025}, as well as interpolation methods \citep{zhu_image_2007, yang_impus_2023}, to the best of our knowledge, no studies explore the effect of using optimal transport to pair prompt embedding tokens for interpolation within a diffusion framework.

In our work, we compare the performance gained by using optimal transport for interpolation between prompt embeddings from a point-cloud view. Although there is evidence that interpolating only between prompt embeddings results in sub-par interpolations \citep{he_aid_2024}, by fixing the latent seed and focusing only on prompt embedding interpolation, we are able to isolate properties and features of the prompt embedding space which otherwise might be masked by interpolating the latents and the prompts simultaneously. Beyond simply improving interpolation methods, this research provides a potential mechanism to improve foundational understanding of the prompt embedding space and emergent behavior. 

In Section \ref{sec:prelims} we present the diffusion and embedding models used, underlying cross attention permutation invariance assumptions required for the application of optimal transport, and the optimal transport methodology. Our novel approach and experimental hypotheses are shared in Section~\ref{sec:approach}. Section \ref{sec:xpmts} establishes our experimental design and presents our results of probing the effect of optimal transport for embedding interpolation. A discussion of the results and potential future work are provided in Section \ref{sec:conclusions}.

\section{Preliminaries and theoretical motivation} \label{sec:prelims}
In this section we present a high-level overview of relevant concepts. We begin with diffusion models, stable diffusion, and  Contrastive Language-Image Pretraining (CLIP) in Section~\ref{subsec:diffu_stab_clip} followed by discussions of CLIP permutation invariance and optimal transport in Sections \ref{subsec:perminv} and \ref{subsec:otwass}, respectively.
\subsection{Diffusion models, stable diffusion, and CLIP} \label{subsec:diffu_stab_clip}

Diffusion models are a class of probabilistic generative models that have garnered significant attention for their ability to produce high-quality images that closely adhere to user-specified criteria, unlocking applications that range from automating artistic content creation to aiding researchers in synthetic data generation, design and modeling complex systems and processes \citep{corso_diffdock_2023, maze_diffusion_2022}. At their core, diffusion models operate through two complementary processes: ({\it i}) a forward ``diffusion'' process where an original image is corrupted through the iterative addition of Gaussian noise, and ({\it ii}) a reverse-diffusion process that reconstructs the original image by estimating and removing the noise added at each diffusion step.  More formally, in the diffusion process, the data $\mathbf{x}_0$ are perturbed at each step $t$ through the Markov chain
\[
q(\mathbf{x}_t | \mathbf{x}_{t-1}) = \mathcal{N}(\mathbf{x}_t ; \sqrt{1-\beta_t} \mathbf{x}_{t-1}, \beta_t \mathbf{I}),
\]
where $\beta_t \in (0, 1)$ is a variance schedule determining the amount of noise added at step $t$, and $\mathcal{N}(\cdot; \mu, \sigma^2)$ is a Gaussian distribution with mean $\mu$ and variance $\sigma^2$.
The reverse-diffusion process to recover $\mathbf{x}_0$ through $\mathbf{x}_T$ is modeled by a separate Markov chain
\[
p_\theta(\mathbf{x}_{t-1} | \mathbf{x}_t) = \mathcal{N}(\mathbf{x}_{t-1} ; \mu_\theta(\mathbf{x}_t, t), \Sigma_\theta(t)),
\]
where $\mu_\theta(\mathbf{x}_t, t)$ is the predicted mean (typically learned using a neural network (U-Net)), and $\Sigma_\theta(t)$ is the variance (often fixed, but can be learned).

While several variations of diffusion models have been developed \citep{cao_survey_2023, ho_denoising_2020, dhariwal_diffusion_2021, rombach_high-resolution_2022, song_score-based_2021, song_denoising_2022}, our interest lies in the widely studied \textit{latent} diffusion model Stable Diffusion 1.5 (SD) by Stability AI \citep{rombach_high-resolution_2022}. Like traditional diffusion models, SD employs a forward and reverse diffusion process, but for SD, these processes operate in a latent space rather than the original data space. A pre-trained variational autoencoder (VAE) first maps input data $\mathbf{x}_0$ to a lower-dimensional latent representation $\mathbf{z}_0$ via an encoder $E$, which substantially improves computational efficiency of the diffusion/reverse-diffusion processes since the latent space is much lower dimensional than that of the original data. The reconstructed representation obtained through the latent reverse-diffusion process is then decoded back into the data space through the VAE decoder $D$. 

The U-Net architecture of the reverse-diffusion process in SD differs from traditional diffusion in that it has been enhanced with attention mechanisms, and in particular cross-attention between the latent space and a textual embedding space.
SD uses the Contrastive Language-Image Pretraining (CLIP) model developed by OpenAI \cite{radford_learning_2021}, which maps images and text strings to a shared embedding space in such a way that paired texts and images are nearby and unrelated pairs are farther apart.
Thus, the Stable Diffusion 1.5 pipeline for producing an image corresponding to a particular prompt string is to compute a CLIP embedding of the string (padded if necessary), which yields a matrix in \(\R^{77\times 768}\).
This is the textual embedding space that is used in the cross-attention blocks in the U-Net architecture for image generation.

In summary, the Stable Diffusion pipeline has two essential ``functions'' around which we will center our focus: ({\it i}) the textual prompt embedding via CLIP and ({\it ii}) a reverse-diffusion processes wherein embeddings are denoised and decoded into an image influenced by the CLIP-encoded text. These two functions can be described as
\begin{itemize}
    \item \(f: \{\text{strings}\} \to \R^{77\times 768}\), where \(s\mapsto f(s)\), the CLIP embedding of the prompt \(s\),
    \item \(g: \R^{77\times 768} \to \{\text{images}\}\), where \(e\mapsto g(e)\), the image generated from embedding \(e\)
\end{itemize}
Note that the image generation function \(g\) also depends on many other parameters, (e.g., the initial ``noisy'' latent), but throughout this paper, we will consider all other parameters fixed. 

\subsection{Permutation invariance} \label{subsec:perminv}

We briefly detail two observations that give rise to an important permutation-invariance property of the function \(g\) above.

The first is a general property about the ubiquitous attention operation. Given two matrices of arbitrary dimension, $X$ and $X^{'}$, a common formulation of the attention operation \citep{vaswani_attention_2017} is 

\begin{equation}\label{eqn:attention}
    A(Q, K, V) = \text{softmax}_{row}\left(\frac{QK^{T}}{\sqrt{D}}\right) V
\end{equation}

where the input matrices are a query ($Q = XW_{Q}$), key ($K = X^{'}W_{K}$), and value ($V = X^{'}W_{V}$) matrix constructed as the product of $X$ or $X^{'}$ and a matrix of learned weights. Equation \eqref{eqn:attention} describes self-attention when \(X = X'\) and cross-attention when \(X\neq X'\). It is a known property that cross-attention is invariant under permutation of the rows of $K$ and $V$, and hence also of \(X'\) \citep{fleuret_deep_2021, ji_mathematical_2019}.

The second critical observation is that the U-Net backbone of Stable Diffusion 1.5 is constructed as a series of self-attention blocks on the image latent and cross-attention blocks between the image latent and the CLIP embedding matrix \citep{rombach_high-resolution_2022}. In the cross-attention blocks, the image latent \(z\) corresponds to $X$ and the CLIP embedding \(e\) corresponds to $X^{'}$.
That is, the function \(g\) defined in \ref{subsec:diffu_stab_clip}, as a function of the CLIP embeddings \(e\), is nothing but a series of cross-attention blocks with \(X' = e\).
Together with the fact that cross-attention is permutation-invariant on the rows of \(X'\), we conclude that the function \(g\) is permutation-invariant on the rows of \(e\).

This means that the \emph{order} of the 77 rows is not relevant, only their values in \(\R^{768}\), which suggests that one can view these embeddings not as \emph{matrices} but instead as \emph{point clouds}.
That is, we can view \(g\) as a function \(g:\cP_{77}^{\text{unif}}(\R^{768}) \to \{\text{images}\}\), where \(\cP_N^{\text{unif}}(\R^d)\) denotes the set of uniform measures on \(N\) (possibly nondistinct) discrete points in \(\R^d\) (i.e., measures of the form \(\frac{1}{N}\sum_{i=1}^N\delta_{x_i}\), where \(x_i\in\R^d\)).
For notational clarity, when viewing embeddings as matrices, we will denote them with Roman letters (e.g., \(e_0\) or \(e_1\)), and when viewing them as point clouds, we will use Greek letters (e.g., \(\mu_0\) or \(\mu_1\)).
Note that there is a many-to-one equivalence between matrices and point clouds, where a matrix \(e\) and point cloud \(\mu\) are equivalent if the rows of \(e\) are exactly the points in \(\mu\).

\subsection{Optimal transport and Wasserstein space} \label{subsec:otwass}

This new point-cloud interpretation of the embedding space comes with a new natural distance metric: the Wasserstein distance.
For two point clouds \(\mu = \sum_{i=1}^N x_i\) and \(\nu = \sum_{i=1}^N y_i\) in \(\cP_{N}^{\text{unif}}(\R^d)\), we define the Wasserstein distance between \(\mu\) and \(\nu\) to be
\begin{equation}\label{eqn:wass-dist-defn}
    W_2(\mu,\nu) \defeq \min_{\sigma\in S_N} \left(\sum_{i=1}^N \|x_i - y_{\sigma(i)}\|_2^2\right)^{1/2},
\end{equation}
where the optimization is over all permutations \(\sigma\in S_N\), the symmetric group on \(N\) elements.\footnote{
    In general, one typically needs to define the Wasserstein distance between discrete measures using the Kantorovich formulation, which allows for the possibility that the optimal ``plan'' may not be induced by a ``map''.
    However, Proposition 2.1 in \cite{peyre_computational_2020} guarantees that, in the case of uniform discrete measures on the same number of points, there exists a permutation that is optimal.
    This is the only case of interest in this paper, so we make the definition of Wasserstein distance using permutations, rather than more general plans.
}
A map \(T\) that satisfies \(T(x_i) = y_{\sigma^*(i)}\) for some optimal permutation \(\sigma^*\) is called an ``optimal transport map'' from \(\mu\) to \(\nu\) and is denoted \(T_\mu^\nu\).
When the support of \(\mu\) is \(N\) \emph{distinct} points (which in practice is almost always true for our CLIP embeddings), then at least one such optimal transport map exists (Proposition 2.1 in \cite{peyre_computational_2020}).
Though not necessarily unique, the notation \(T_\mu^\nu\) will refer to a particular choice of an optimal map, which we will call ``the'' optimal transport map.

While we lose the Euclidean space structure that comes with the matrix perspective, the point-cloud perspective comes with an analogous mathematical structure of a (formal) Riemannian manifold \citep{ambrosio_gradient_2008, otto_geometry_2001}
Specifically, \(W_2\) is a metric on \(\cP_N^{\text{unif}}(\R^d)\), and \(\cP_N^{\text{unif}}(\R^d)\) is a subset of the ``Wasserstein manifold'' of measures with finite second moment, which itself comes with some useful geometric properties.
As we begin to explore the landscape of embedding space with this new perspective, the key geometric property of interest to this paper is the nice relationship between barycenters and geodesics.
Specifically, if \(T_{\mu_0}^{\mu_1}\) is an optimal transport map between \(\mu_0,\mu_1 \in \cP_N^{\text{unif}}(\R^d)\), then there is an explicit expression for a ``constant-speed geodesic'' \(\mu:[0,1]\to \cP_N^{\text{unif}}(\R^d)\) from \(\mu_0\) to \(\mu_1\) given by \(t\mapsto \mu_t\) with
\begin{equation}\label{eqn:wass-geod}
    \mu_t = ((1-t)\id + tT_{\mu_0}^{\mu_1})_\#\mu_0 = \frac{1}{N}\sum_{i=1}^N \delta_{(1-t)x_i + ty_{\sigma^*(i)}},
\end{equation}
where the notation \(T_\# \mu\) denotes the pushforward of \(\mu\) by the map \(T:\R^d\to\R^d\) \citep{ambrosio_gradient_2008}.
In particular, for \(t\in[0,1]\), \(\mu_t\) is the weighted Wasserstein barycenter; i.e., \(\mu_t\) satisfies
\begin{equation}\label{eqn:wass-bary}
    \mu_t = \arg\min_{\mu\in\cP_N^{\text{unif}}(\R^d)}\left[(1-t)W_2(\mu_0,\mu)^2 + t W_2(\mu_1,\mu)^2\right].
\end{equation}
This barycenter definition is exactly analogous to a ``weighted average'' in Euclidean space, which means these geodesics between point clouds on the Wasserstein manifold are exactly analogous to straight lines --- they trace out the shortest path between two points in the space.
Thus, in essense, this says that the OT-inspired way to interpolate two \(N\)-point clouds \(\mu_0, \mu_1\) is to pair off points in \(\mu_0\) with points in \(\mu_1\) using the optimal transport ``coupling'' (i.e., the pairing of \(x_i\) with \(T_{\mu_0}^{\mu_1}(x_i) = y_{\sigma^*(i)}\)) and then to linearly interpolate between each pair simultaneously.
Despite losing the nice structure of the vector space of matrices, the point-cloud interpretation of the embedding space still exhibits a natural way to interpolate between embeddings. Exploiting the point-cloud interpretation of the learned embedding space provides the basis for our novel, geometrically-informed image interpolation technique presented in Section \ref{sec:approach}

\section{Geometry-informed Image Interpolation Approach}
\label{sec:approach}

In this work we explore how the permutation-invariance property described in Section \ref{subsec:perminv} can inform novel image interpolation techniques by operating on the embeddings.
Specifically, the discussion above suggests that the natural way to interpolate between embeddings (viewed as point clouds) is to use \eqref{eqn:wass-geod} with the optimal coupling \(\sigma^*\).
This gives the shortest path through the embedding space between the two prompts, meaning that any other method of interpolating the embeddings gives a longer (or at least not shorter) path through Wasserstein space.
For example, there is a natural way to interpolate between embeddings \(e_0, e_1\) viewed as matrices, which is to linearly interpolate by setting \(e_t = (1-t)e_0 + te_1\).
By viewing \(e_t\) now as a point cloud, this method traces out a path through Wasserstein space also described by \eqref{eqn:wass-geod}, except instead of using the optimal coupling \(\sigma^*\), it uses the coupling induced by the order of the rows in the CLIP matrices, which we call the ``CLIP coupling.''
In fact, the length of this path is exactly the cost of the associated coupling (consequence of Theorem 8.3.1 in \cite{ambrosio_gradient_2008}), and for the CLIP coupling, that cost is the standard 2-norm of the difference of corresponding rows. 
Thus, not only do we know that the CLIP interpolating path is longer (since the optimal coupling cost is at least as small as the CLIP coupling cost), we know precisely how much longer.

One way to assess whether our point-cloud perspective on embedding space is ``more natural'' than the matrix perspective is to investigate how these path lengths through embedding space relate to notions of similarity in image space.
Specifically, for a particular interpolation path through embedding space, there is an associated path through image space obtained by generating the image that corresponds to each interpolated embedding along the embedding path. 
For any embedding-interpolation path, the start and end embeddings are fixed, and this means that the associated image path connects the images corresponding to the start and end embeddings.
If one embedding-interpolation path is ``better'' than another, its associated image-interpolation path should be smoother, more natural-looking, and contain images which are more similar.
In short, the path through image space should be ``shorter.''
Thus, if considering the CLIP embeddings as point clouds really is a more faithful geometric interpretation of embedding space, we hypothesize that:
\begin{enumerate}
    \item The geodesic path through embedding space is shorter for prompts which are more similar (i.e., embeddings are closer in Wasserstein distance when the prompts are more similar), and the associated image paths have lower PPL scores,
    \item suboptimal couplings give relatively worse embedding interpolations for prompts which are more similar (because suboptimal couplings are relatively more costly when embeddings are close in Wasserstein distance)
    \item for a fixed pair of prompt embeddings, a shorter interpolating path between them gives a better associated image interpolation, and
    \item this effect increases for prompts which are more similar because the relative path-length increase is less severe.
\end{enumerate}

To properly test these hypotheses, we need a suitable notion of path length in image space.
Rather than use a pixel-wise metric, which promotes structurally smooth but unrealistic image interpolations, we seek an image-similarity score that promotes smooth image transitions while retaining realism and context shifts.
To this end, we use Perceptual Path Length (PPL) \cite{karras_style-based_2019} as a surrogate for path length.
PPL is defined as the average of image similarity scores between consecutive equally-spaced images along a path, with the intuition that equally spaced points are closer if the overall path is shorter.
The standard image similarity score used for PPL is Learned Perceptual Image Patch Similarity (LPIPS) \cite{zhang_unreasonable_2018}, which is a standard method for judging image similarity.

\begin{figure}[h!]
  \centering
  \includegraphics[width=1\textwidth]{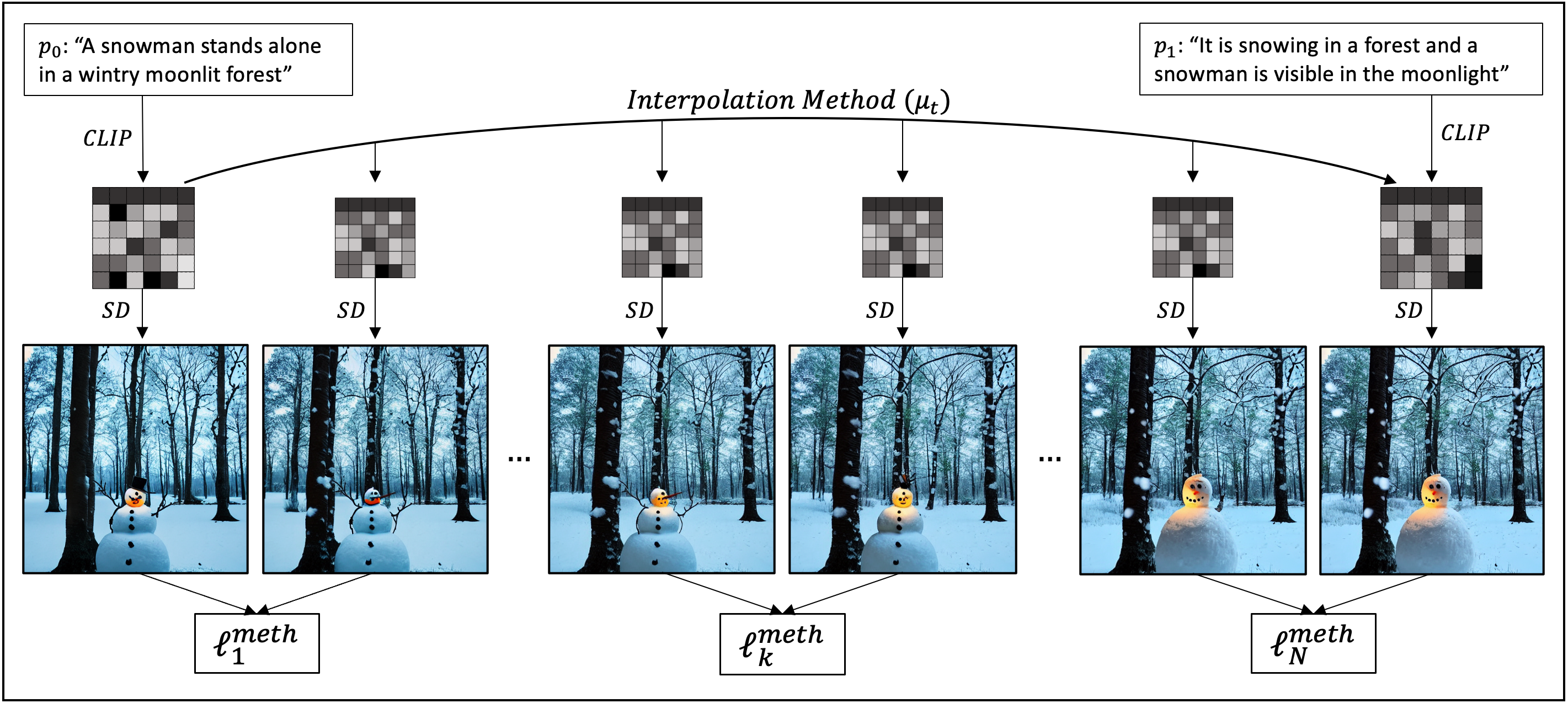}
  \caption{Embedding interpolation workflow where SD indicates use of the Stable Diffusion model to produce the associated image.}  \label{fig:diagram}
\end{figure}

Thus, our high-level approach is as follows, as illustrated in Figure \ref{fig:diagram}.
Two embedding matrices are obtained from two prompt pairs using CLIP.
The embedding matrices are treated as point clouds, and three interpolation couplings (OT, CLIP, and a random coupling) are used to define a path between the embedding point clouds.
The interpolated path in embedding space is sampled along a grid, and resultant embedding point clouds are obtained.
Both the original embeddings and the interpolated embeddings are used to produce a corresponding trajectory of images in image space.
For each grid sample along the interpolated path in embedding space, we produce a corresponding image with a diffusion model (Stable Diffusion), and the resultant images are compared using PPL, based on the LPIPS scores between the $k^{th}$ and $(k+1)^{th}$ image denoted $\ell_{k}$.
In Section \ref{sec:xpmts}, we present experimental results that use this pipeline to explore the validity of the hypotheses listed above.

\section{Experimental Design and Results} \label{sec:xpmts}

\subsection{Experimental setup} \label{subsec:xpmt-setup}

To evaluate the impact of capturing the geometric structure of CLIP embeddings in image interpolation, we consider many pairs of prompts (see below for more details about how the prompt pairs were selected).
For each pair, we follow the pipeline illustrated in Figure \ref{fig:diagram} for each of three couplings --- OT, CLIP, and a random coupling (to create the random coupling, we fix the first rows of the CLIP matrices --- this row is always the same for any prompt --- and randomly pair the last 76 rows in each CLIP matrix).
We compute the cost of each coupling and the resulting PPL score for each coupling's image interpolation.
We note that LPIPS gives lower scores to images that are more similar, and so a smaller PPL score means that the given path through image space is ``shorter''.
Since the OT interpolation method produces a geodesic \(\mu_t^{\text{OT}}\) through embedding space, and each of the other two interpolations \(\mu_t^{\text{clip}}\) and \(\mu_t^{\text{rand}}\) are longer paths, we hypothesize that the LPIPS/PPL scores for the OT method \(\ell_k^{\text{ot}}\) will be lower on average than the other methods.

\subsection{Dataset}

To properly test the efficacy of the three interpolation methods, we must produce interpolated images from each method for many pairs.
The structure of our hypothesis requires a dataset with prompt pairs that carry a known similarity score. To create such a dataset, we leverage the Crisscrossed Captions dataset \citep{parekh_crisscrossed_2021}, which contains captioned images from the the MS-COCO dataset \citep{lin_microsoft_2015} that have been manually scored according to their similarity on a scale from 0 to 5, with 5 being most similar. We assume that the similarity score between two images can be extended as a similarity score the associated pair of captions. We discretize the range of similarity scores into bins of width 0.5 and randomly select 1,000 pairs from each similarity group (excluding any pairs where the two captions were identical). These captions are then used as prompts in the SD model with a fixed seed. The results shown reflect the experiment described in Section \ref{subsec:xpmt-setup} applied to 10,000 caption/prompt pairs. The Crisscrossed Captions dataset has a custom, open source license at \href{https://github.com/google-research-datasets/Crisscrossed-Captions/blob/master/LICENSE}{https://github.com/google-research-datasets/Crisscrossed-Captions/blob/master/LICENSE}. The MS-COCO dataset has a Creative Commons Attribution 4.0 License, and the SD model has a CreativeML Open RAIL-M License. CPU and GPU workers were used on an internal cluster and the total compute cost for the experiments was $1,094$ CPU hours and $2,172$ GPU hours. Preliminary and failed experiments accounted for an additional approximately $7,000$ CPU hours and $575$ GPU hours of compute resources.

\subsection{Results} \label{subsec:xpmt_res}
\begin{figure}[h!]

  \centering
  \includegraphics[width=1\textwidth]{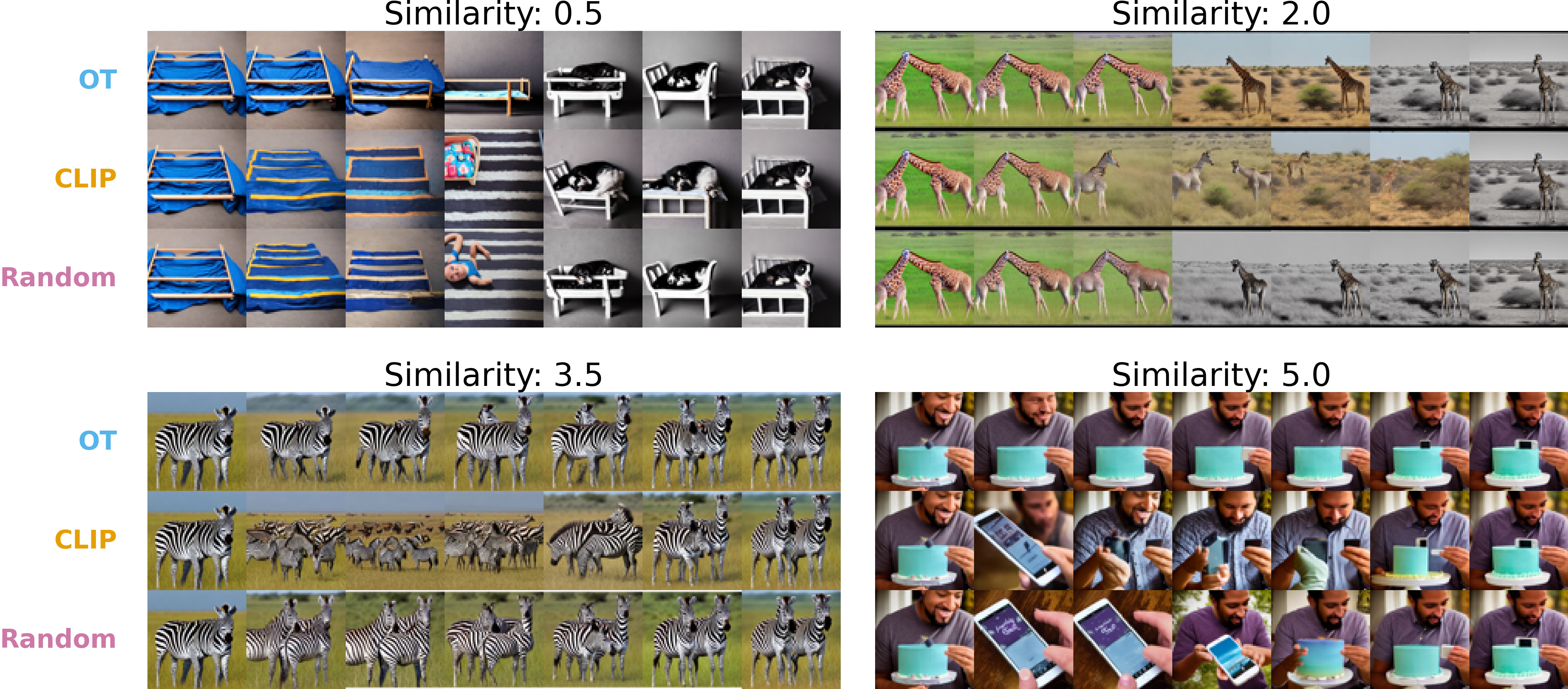}
  \caption{Image interpolations for each method across four selected prompt pairs of varying similarity.}  \label{fig:arrayplots}
\end{figure}

To help build the reader's intuition, we start by showing the qualitative impact of the geometric relationships being considered. Figure~\ref{fig:arrayplots} shows a selection of interpolated images, where for each of the four panels, the sub-images on the far left and far right correspond to the images produced for each prompt pair, i.e., the endpoints of our path in image space. For each panel, the rows of sub-figures are ordered from top to bottom as OT, CLIP, and random. Examination of the image trajectories shows cases where the CLIP and/or random method appears to hallucinate objects in the interpolated trajectory. The prompt pairs shown were selected based on having quantitatively larger differences in the associated path lengths for illustration purposes. However, as will be shown subsequently, the OT path is found to perform comparably or above the CLIP or random couplings across the board.

For a quantitative evaluation, we compute the perceptual path length (PPL) (as defined above in Section~\ref{subsec:xpmt-setup}) \citep{karras_style-based_2019} for each interpolation method (OT, CLIP, random) as a measurement of transition smoothness between the trajectory of interpolated images for all 10,000 pairs of prompts.  
Smaller PPL values are desirable and correspond to higher smoothness across the interpolated image trajectories. In the subsequent plots, we report PPL and coupling costs over the collection of all 1000 prompt pairs of each similarity level. 

\begin{figure}[h!]

  \centering
  \includegraphics[width=1\textwidth]{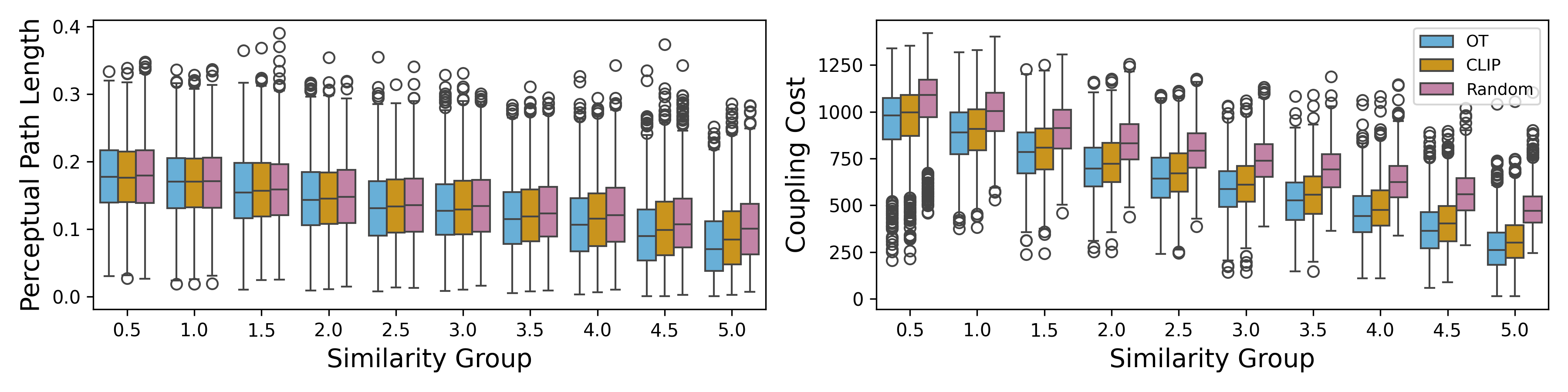}
  \caption{PPL (left) and coupling cost (right) by interpolation method and similarity group.}  \label{fig:boxplots}
\end{figure}

The boxplots of PPL scores and coupling costs are plotted for each method and similarity group in Figure~\ref{fig:boxplots}. To test our hypothesis, we assess the statistical significance of both the difference in median PPL and the difference in median coupling cost between the OT interpolation results and the CLIP or random couplings, respectively within each similarity group. Significance of p-values for testing the difference in median PPL within each interpolation method and similarity group are reported in Table \ref{table:pvalues_sig}. As hypothesized, the impact of embedding path is more significant for more similar pairs of prompts. All tests for the difference in median coupling cost between optimal transport and each other interpolation method are highly significantly different, $p<0.0001$, for all similarity groups. Additionally, the path length decreases as a function of similarity and the significance of an optimal coupling grows as the similarity increases as hypothesized based on geometric intuition. The PPL scores and coupling costs were determined to be not normally distributed, so for both sets of tests we employed a Wilcoxon signed-rank test to test the null hypothesis that the median of the difference is zero. In addition to the typical assumptions of independence and randomness, the Wilcoxon signed-rank test assumes continuous data distributed symmetrically about the median \citep{ramachandran_chapter_2021}.

\begin{table}[h!]
  \caption{Significance of p-values for the Wilcoxon test comparing the median of the PPL scores. \\ ($p < 0.05^{*}$, $p < 0.01^{**}$, $p < 0.001^{***}$)}
    
  \label{table:pvalues_sig}
  \centering
\begin{tabular}{lcccccccccc}
    \toprule
    & \multicolumn{10}{c}{Similarity Group} \\
    & 0.5 & 1.0 & 1.5 & 2.0 & 2.5 & 3.0 & 3.5 & 4.0 & 4.5 & 5.0 \\
    \cmidrule(r){2-11}
    OT vs. CLIP PPL  & -&	-	&-&	-&	$***$	&$*$	&$***$	&$***$	&$***$	&$***$ \\
    OT vs. Random PPL & $*$&	-	&$**$&	$***$&	$***$	&$***$	&$***$	&$***$&$***$&	$***$ \\ 
    \bottomrule
  \end{tabular}
\end{table}

\section{Conclusions and Future Directions}\label{sec:conclusions}

In this work we observe that interpolating using the OT coupling in general results in shorter image paths than using the linear and random couplings, and that the improvement is more substantial for more similar prompts. Both of these observations are consistent with the hypothesis that path length through Wasserstein space is reflected by ``path length'' through image space.
We also notice that in many cases where OT drastically outperforms the other two methods, it does so because the other methods produce interpolated images which are relatively very different than either end image.
This suggests that embedding space has a ``convexity'' property with respect to Wasserstein distance that it does not have when the embedding matrices are handled as matrix objects. Concretely, we find given two embeddings of a particular class, an embedding ``between'' them in Wasserstein space is more likely to be in the same class than an embedding ``between'' them in matrix space.
Our experimental results suggest that the point-cloud perspective indeed does a better job of capturing the ``geometry'' of embedding space in ways that reflect more desirable properties in image interpolation. 

Deeper exploration of the optimal path between prompt embeddings could uncover properties and conditions under which prompt interpolation results in thematically consistent and smooth images, giving rise not only to improved methods for image interpolation that rely on prompt interpolation \citep{wang_interpolating_2023}, but also informing scenarios where prompt interpolation or manipulation is an auxiliary step, such as variant refinement \citep{deckers_manipulating_2024}. An area for future work is in developing additional metrics to capture pair-wise changes in images. The path length surrogate considered in this work leverages a standard image similarity score, LPIPS, but that score does not explicitly capture or penalize for contextual and content shifts between image pairs. Additional work in this area would enable more direct connections with applications like hallucination detection and associated mitigation strategies.

Finally, the scope of this work focused solely on Stable Diffusion 1.5, whose architecture differs substantially from the latest Stable Diffusion model (3.5) available at the time of writing. While architectural innovations in the latest models present barriers to direct extension of the proposed approach, the point cloud perspective may still better preserve underlying geometry and desired invariances than performing operations like token concatenations used in Stable Diffusion 3.5. Therefore, an important future direction would be extending and evaluating the point cloud framing for shared, multi-modal token spaces and transformer-based denoising architectures like those of current state-of-the-art models.

\begin{ack}   
This work was conducted under the Laboratory Directed Research and Development Program at PNNL, a multi-program national laboratory operated by Battelle for the U.S. Department of Energy under contract DE-AC05-76RL01830.
\end{ack}

\bibliographystyle{apalike}

\bibliography{tagds_ddot_zotero_bib}

\end{document}